\documentclass[letterpaper, 10 pt, conference]{ieeeconf}  

\IEEEoverridecommandlockouts                              

\usepackage{xcolor}
\usepackage{amsmath} 
\usepackage{graphicx}
\usepackage{subcaption}
\usepackage{booktabs}
\usepackage{float}
\usepackage{placeins}

\title{\LARGE \bf
Value Function-Guided MPPI for Hierarchical Planning and Control in Off-Road Autonomous Navigation
}

\author{Anoushka Alavilli$^{1}$, Raymond Song$^{1}$, Guanya Shi$^{1}$, and Jeff Schneider$^{1}$
\thanks{*This work was not supported by any organization}
\thanks{$^{1}$The authors are with the Robotics Institute,
Carnegie Mellon University, Pittsburgh, PA 15213, USA.
{\tt\small \{apalavil, rysong, guanyas, jeff4\}@andrew.cmu.edu}}%
}

\begin{document}

\maketitle
\thispagestyle{empty}
\pagestyle{empty}

\begin{abstract}

Autonomous navigation in off-road terrain is commonly achieved using hierarchical control and planning architectures that rely on explicit cost maps and one-way information flow from a high-level planner to a low-level controller, decoupling planning from the true execution cost experienced during control. We propose a hierarchical planning and control framework in which information flows bidirectionally, allowing the low-level controller to guide trajectory selection at the planning level. The highest level of the hierarchy is a Model Predictive Path Integral (MPPI) control-style planner that determines optimal trajectories and passes these waypoints to a pretrained low-level reinforcement learning (RL) controller. Rather than drawing from an explicit cost map, our MPPI planner leverages the value function of the RL controller as an implicit cost representation. In this formulation, MPPI serves as a trajectory exploration mechanism, while the low-level controller’s value function guides the search. To account for the distribution shift seen by the value function between training and evaluation, we propose an online policy evaluation process that adapts the value function to unseen driving experiences, and we demonstrate this adaptation in unseen off-road simulation environments. We evaluate our approach in a high-fidelity off-road driving simulator across a range of challenging terrain conditions and compare its performance against a baseline MPPI controller. The results demonstrate that our method achieves safer and more effective navigation to goal locations while eliminating the need for explicit cost maps and enabling tighter coupling between planning and control.

\end{abstract}

\section{INTRODUCTION}

\begin{figure*}[t]
    \vspace{2mm}
    \centering
    \includegraphics[width=0.9\textwidth]{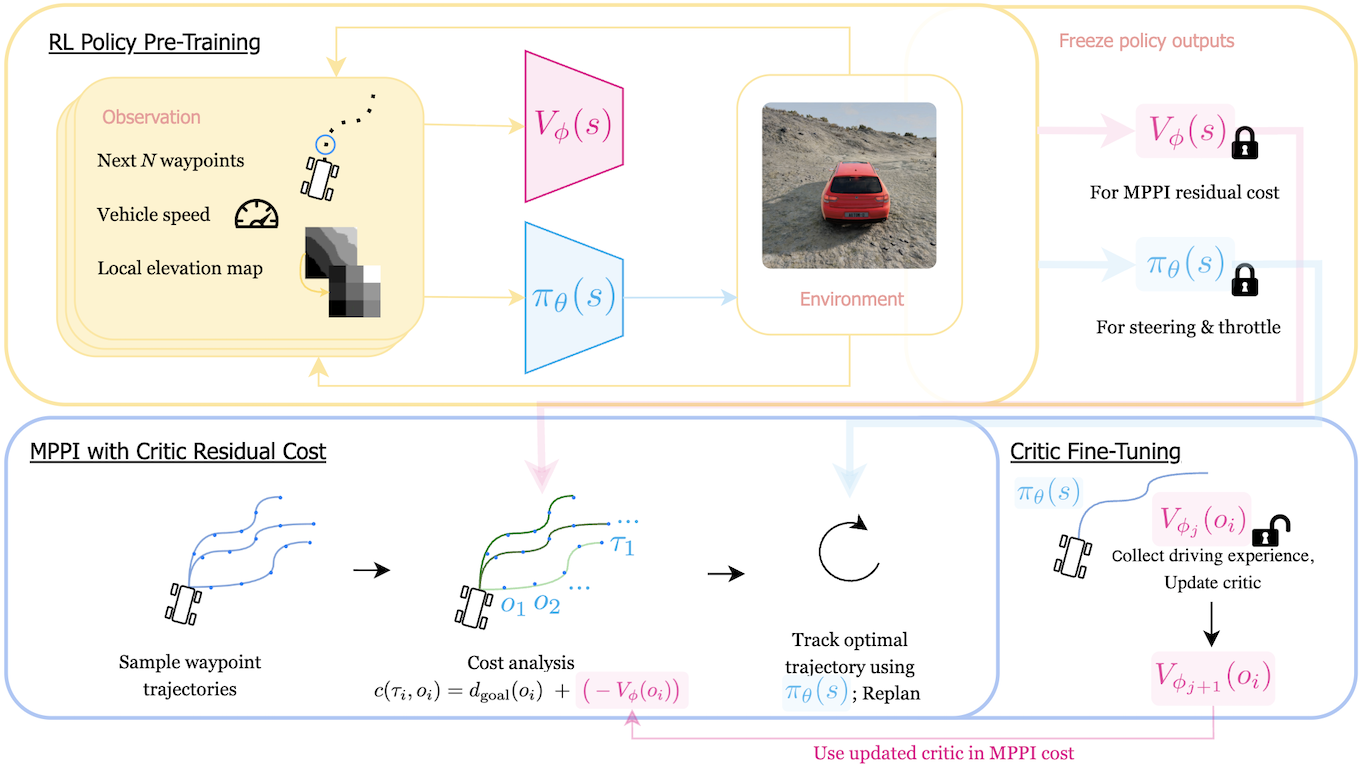}
    \caption{Overview of our value-guided planning framework. We pre-train RL policies to provide both an actor $\pi_\theta(s)$ and a value function $V_\phi(s)$. MPPI samples candidate trajectories and evaluates them using a cost that combines goal progress with the negative value estimate, treating $-V_\phi(s)$ as a control-aware cost. The selected trajectory is executed by the policy, and the resulting experience can be used to update the value function online. This enables planning to be guided by the expected performance of the controller without requiring an explicit cost map.}
    \label{fig:algorithm}
\end{figure*}

Autonomous navigation in off-road environments requires vehicles to safely traverse from a start to a goal location while interacting with complex, unstructured terrain. This task is challenging due to both terrain properties, such as low traction in sand or deformability of mud, and terrain geometries, including steep slopes, ditches, and uneven surfaces that can induce unsafe vehicle dynamics such as excessive pitch and roll. While there is substantial prior work on perceiving and modeling both categories of terrain features, in this work we focus specifically on planning and control under challenging terrain geometries. Our goal is to enable a vehicle to safely navigate from start to goal by selecting and executing trajectories that avoid or carefully traverse geometrically hazardous regions.

Central to this work is the notion that the true “cost” of traversing terrain is highly dependent on the control that is deployed. For example, take the situation of a vehicle driving down a steep slope. The vehicle might experience unsafe toppling if it passes over the slope quickly, but it could also feasibly pass over the slope safely at a lower speed. Therefore, the “cost” of this hill is not only dependent on the hill geometry but also on the control law deployed. Results from works such as \cite{howdoesitfeel} validate this hypothesis in their findings that velocity observations aid in predicting costs of trajectory traversal.

Model Predictive Path Integral (MPPI) control \cite{mppi_original_paper} is a powerful method that has had much demonstrated success in autonomous locomotion, both in off-road and agile locomotion tasks \cite{han2024mppi, gibson2023multistep, roadrunner, roth2025fdm, wholebodymppialvarez}. Key to MPPI’s success are the parallelizable open-loop dynamics rollouts, the optimization costs for which are generally determined from a learned or predefined cost map of the terrain. These cost maps can be learned via methods such as semantic and geometric assessment \cite{bevnet} or via self-supervised learning \cite{terrainnet, v-strong}. Simultaneously, reinforcement learning (RL) policies have demonstrated strong performance in learning control laws for agile robots \cite{ppo, Heess2017EmergenceOL, learningagilemotorskills, learningtowalk}. Unlike traditional cost maps, which assign costs based primarily on terrain understanding, RL policies learn value functions that estimate expected return under the control behavior they induce. Consequently, these value functions provide an implicit estimate of trajectory cost conditioned on the control policy. This raises a natural question: rather than relying on an explicit cost map, can we instead use a policy’s learned value function as a measure of terrain drivability for navigation?

In traditional hierarchical autonomy stacks, information flows top-down: a high-level planner proposes trajectories, and a low-level controller attempts to track these trajectories, as in \cite{valueofplanninghatch}. This decoupling can lead to repeated failures when the planned trajectories are not feasible under the true execution dynamics of the controller, and particularly in off-road environments where terrain-vehicle interactions are complex and highly nonlinear \cite{gibson2023multistep}. In our hierarchical algorithm, the low-level controller actively informs the planning process. By using the value function of the low-level RL controller to provide an estimate of the expected return under its own control policy, the planner naturally selects trajectories that the robot is expected to be able to traverse. This enables the planner to favor trajectories that not only meet navigation goal requirements but also are realizable by the low-level tracking controller.

Our hierarchical algorithm effectively induces an implicit cost map, shaped by the controller’s learned understanding of terrain difficulty and vehicle limitations, without requiring explicit terrain labeling or cost map learning. By leveraging the value function as an execution-aware cost signal, we enable planning to account for the coupled effects of terrain and control without explicitly modeling them. This perspective frames cost maps not as standalone learned representations of the environment but as quantities that can be implicitly derived from the capabilities and limitations of the control policy itself. 

We summarize our main contributions as follows:
\begin{itemize}
    \item The first formulation of MPPI that uses, as a cost function, the negative expected return estimated by a pretrained RL value function.
    \item A hierarchical planning framework with bidirectional feedback between planning and control, enabling online adaptation that mitigates repeated control failures.
    \item An off-road navigation algorithm that can plug into a real autonomy stack with ease, as the algorithm does not use privileged information from the simulator that cannot be adequately replicated using sensory inputs from onboard systems.
    \item A new perspective on the form and role of cost maps for off-road driving.
\end{itemize}

\subsection{Costmap Learning, Traversability Estimation, and Dynamics Modeling}

Prior work in off-road navigation has focused on learning terrain representations for planning and control, typically through costmaps, traversability estimation, or learned dynamics models. These methods differ in their observation modality, how they are trained, and in the type of information they aim to predict.

\textbf{Explicit Terrain Representations.} Approaches such as \cite{bevnet} and \cite{terrainnet} learn dense bird’s-eye view (BEV) representations from sensor data using offline supervision. \cite{bevnet} constructs traversability costmaps from sparse LiDAR and uses inpainting for occluded regions. \cite{terrainnet} extends this approach to camera-based perception, predicting multi-layer terrain representations for both ground and ceiling structure. While \cite{terrainnet} incorporates self-supervised depth completion and pseudo-labeling, it relies on manually annotated data. These BEV maps integrate well with planning but require curated datasets and do not adapt online.

Recent work reduces reliance on manual labels by leveraging self-supervision from onboard sensing and interaction. \cite{roadrunner} learns BEV elevation and traversability maps using hindsight labels, while works such as \cite{badgr} learn terrain affordances directly from real-world experience. \cite{alter} extends this to online adaptation by continuously updating a visual traversability model using LiDAR supervision. Other approaches include using proprioceptive feedback to aid in visual understanding \cite{Chen24physical}, or contrastive learning for traversability estimation \cite{v-strong}.  Uncertainty-aware traversability estimation has also been explored \cite{step, alter}. While these methods improve scalability and generalization, they typically produce static cost predictions not explicitly conditioned on the controller. Some works \cite{speedmap} partly account for this gap directly by predicting traversal speed in their cost maps.

Our approach assumes that 3D elevation data is sufficient for estimating terrain traversability. In practice, terrain semantics may provide additional information about properties such as deformability (e.g., mud or sand). In this work, however, we focus on the planning–control interface rather than perception. Our framework is compatible with richer perceptual inputs, which could be incorporated into the RL policy observations without changing the overall algorithm.

\textbf{Learning Costmaps from Demonstration and Interaction.} In \cite{airlab_irl}, an explicit discretized costmap is learned by minimizing the discrepancy between expert and MPPI-generated state visitation distributions, favoring routes from MPPI that are similar to those taken by the expert. WVN \cite{fast_wild_visual_nav_frey}, which addresses traversability for quadrupedal robots, also uses human trajectory data but in a different manner. The robot learns from a few minutes of human data and correlates poor velocity tracking with lower traversability scores. After the inital data collection phase, the robot navigates by itself using its learned costmap. Thus, both of these approaches learn where the robot should travel, whereas ours jointly considers the terrain and the control policy that influences \textit{how} robot executes the route.

SALON \cite{salon} uses self-supervision to associate BEV terrain features with costs measured during prior traversal. It learns a velocity-conditioned costmap that predicts expected ``roughness" (quantified by IMU and suspension readings) and a speedmap that estimates the maximum speed allowable under a user-specified roughness limit. Similar to SALON, METAVerse \cite{metaverse} also uses self-supervised learning from physical vehicle-terrain interactions to continuously adapt cost predictions in real time. METAVerse associates LiDAR geometry and speed with traversability cost quantified by vertical acceleration and then adapts a neural costmap predictor using gradient updates from recent experience. Our work is similar in motivation to both of these works; however, our MPPI cost formulation captures a tradeoff between terrain difficulty and goal progress. By training the value function with a reward that jointly considers progress as well as controller and terrain penalties, our planner may select a rougher yet feasible route for goal-reaching tasks.

\textbf{Learned Dynamics for Planning.} An adjacent line of work learns predictive models of vehicle dynamics for use in planning. Methods such as \cite{roth2025fdm} predict dynamics from proprioception and local terrain geometry as part of their MPPI framework. Other works, such as \cite{gibson2023multistep}, rely strictly on odometry and proprioception by using a ``hybrid" architecture. Works such as \cite{dinov2alavilli} improve upon this category of approach by incorporating visual terrain information in the dynamics model predictions to better capture terrain interactions. These methods implicitly capture terrain difficulty and are useful for predictions over short horizons while cost maps serve as a method for longer-horizon cost analysis.

\textbf{Online Adaptation of Learned Models.} Several works address the challenge of adapting to novel or out-of-distribution environments through online learning. In perception, methods such as \cite{alter} and \cite{Chen24physical} use self-supervision to continuously update visual models during deployment. In contrast, other approaches such as \cite{metadynamics} adapt the control model itself to allow model-based controllers such as MPPI to better reflect the current terrain conditions. In our method, we adapt the value function of our RL policy to better predict the costs experienced during deployment.

Overall, prior work either learns explicit traversability maps or short-horizon dynamics models to assess costs. In contrast, our approach does not learn a model and instead uses a learned value function to provide a control-aware estimate of terrain difficulty that directly shapes trajectory optimization, bridging control and planning without requiring explicit costmap supervision or dynamics modeling.

\subsection{Value Functions for Planning and Control}

Recent work has explored incorporating value functions into model predictive control (MPC) to mitigate the myopic behavior induced by short planning horizons. TD-MPC \cite{tdmpc} learns a parametric Q-value function via temporal difference learning alongside a latent dynamics model, and uses this value function to estimate the terminal return of sampled trajectories. To assess the costs of non-terminal waypoints, TD-MPC uses a jointly-learned reward function that is separate from the value function. In contrast, \cite{valueofplanninghatch} proposes a non-parametric alternative by treating a sampling-based planner as an approximate value function, where each node encodes a cost-to-go estimate that can be queried during control. Both approaches use value functions to bootstrap the terminal cost of trajectories, improving long-horizon reasoning without increasing planning complexity. However, these methods either rely on learned reward models and extensive training \cite{tdmpc} or assume access to a global planning structure or map \cite{valueofplanninghatch}, which may not be available in unseen environments. In contrast, our approach uses a learned value function not only as a terminal bootstrap but also as a dense, waypoint-wise cost that shapes the trajectory over a local horizon rather than across a global map.

\begin{figure}[t]
    \vspace{2mm}
    \centering
    \includegraphics[width=0.95\linewidth]{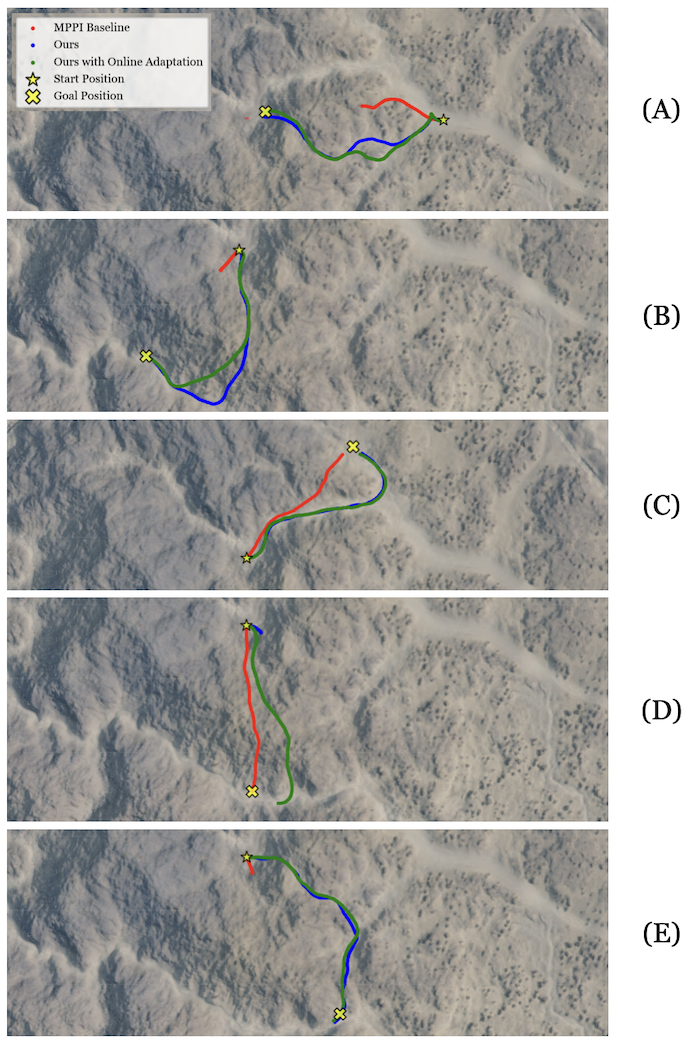}
    \caption{Comparison between baseline MPPI, our algorithm, and our algorithm with online adaptation. All trajectories were unseen during training. In (A), our method incurs damage midway through the trajectory, while the online adaptation variant avoids this region through online updates alone, without any prior knowledge that this region would cause damage. In (A), (B), and (E), the baseline MPPI algorithm becomes stuck and is unable to complete its planned route. In (C), our algorithm demonstrates its ability to follow existing trails, which provide safer traversal than off-trail terrain. In (D), our algorithm with online adaptation reaches the maximum time limit and does not complete the trajectory.}
    \label{fig:traj_comparison}
\end{figure}

\section{METHODS}\label{sec:methods}

\subsection{RL Policy Training}\label{sec:methods:ppo}
We first train a low-level RL controller to track waypoint sequences. All training and evaluations are conducted in the Johnson Valley environment in BeamNG ~\cite{beamng}. The controller is trained using Proximal Policy Optimization (PPO) ~\cite{ppo}. Training waypoint trajectories are constructed from over $3.5$K meters of human-teleoperated driving trajectories collected in the simulator. At each timestep, the policy receives as input the next waypoints along these trajectories and is trained to track them. The learned policy outputs continuous control commands (steering and throttle) to follow the waypoint sequence.

The policy observation consists of (i) the distance and heading to the next $N=3$ waypoints, (ii) the vehicle’s current speed, and (iii) a $6 \times 6$ meter elevation map of the terrain directly in front of the vehicle, downsampled to $3 \times 3$. All observations are stacked over the previous three frames. The elevation map is derived from a nadir-pointing depth camera in the simulation and post-processed into a local height map before being provided to the policy. We use only geometric terrain information and do not include semantic labels or other visual features. Because the map is generated directly from simulator depth measurements, it can be viewed as a ground-truth elevation observation and contains no unknown cells. Vegetation and rocks are present in the initial raw depth reading; however, because the elevation map is heavily downsampled, the effects of vegetation and small rocks are largely removed from the final height-map observation in practice. Importantly, vegetation does not alter the dynamics of the vehicle in the simulator. The reward function (Eq. \ref{eq:reward_fn}) penalizes unsafe behavior, including damage, vertical acceleration, and vertical jerk, while rewarding progress toward waypoints, as described by:

\begin{equation}
\label{eq:reward_fn}
\begin{aligned}
r(s,a) =\;& \beta_1 \Delta \text{waypt dist} + \beta_2 \Delta \text{damage} \\
          &+ \beta_3 \text{clip}(\text{jerk}_z) + \beta_4 \text{clip}(\text{acceleration}_z),
\end{aligned}
\end{equation}

where $\beta_1, \beta_2, \beta_3, \beta_4$ are reward weights defined in the Appendix. 

To integrate policy value estimates into planning, we additionally train a separate critic on single-waypoint tracking tasks. This critic observes only the next waypoint ($N=1$) and is trained on episodes that terminate once the vehicle reaches within a distance $d_{\text{acceptance threshold}}$ of the waypoint. The training is otherwise unchanged from that of the waypoint-tracking policy. This design isolates the cost of executing short trajectory segments, enabling consistent evaluation of candidate trajectories within the MPPI framework. While a single-waypoint policy could also be used for control, we find it insufficient for stable driving at higher speeds due to the lack of lookahead information. See Fig. \ref{fig:algorithm} for a summary of the policy and critic training as well as the Appendix for a complete list of all hyperparameters used in training.

\subsection{MPPI with Critic Residual Cost}\label{sec:methods:mppi}

The second level of our hierarchical algorithm is an MPPI-style planner that incorporates the pre-trained critic (Section \ref{sec:methods:ppo}) as a residual cost. The planner samples trajectories in action space (steering and throttle) and propagates them forward using a kinematic bicycle model with Ackermann steering to produce candidate waypoint trajectories.

The cost of each trajectory (Eq. \ref{eq:mppi_traj_cost}) is computed as the sum of waypoint-wise costs along the trajectory. Each waypoint cost (Eq. \ref{eq:mppi_waypt_cost}) consists of (i) a Euclidean distance term that encourages progress toward the goal and (ii) a residual cost given by the negative expected return estimated by the pre-trained critic. This residual term provides a control-aware measure of terrain difficulty. To ensure goal completion, the residual cost weight $\lambda_{\text{res}}$ is set to zero when the vehicle is within one planning horizon of the goal, as described by Eq. \ref{eq:mppi_alpha}. Let $\tau=\{w_i\}_{i=1}^{N}$ denote a candidate trajectory containing
$N$ waypoints, $o_i$ denote the critic observation constructed at
waypoint position $w_i$, $g$ denote the goal position, and
$d_{\text{horizon}}$ denote the planner's spatial horizon, 40 m.

\begin{equation}
\label{eq:mppi_traj_cost}
C(\tau) \;=\; \sum_{i=1}^{N} c(w_i, o_i),
\end{equation}

\begin{equation}
\label{eq:mppi_waypt_cost}
c(w_i, o_i)
\;=\;
\lambda_{\text{goal}} \,\lVert w_i - g \rVert_2
\;+\;
\lambda_{\text{res}} \,\alpha_i \,\big(-V_{\text{ppo}}(o_i)\big).
\end{equation}

\begin{equation}
\label{eq:mppi_alpha}
\alpha_i \;=\;
\begin{cases}
0, & \text{if } \lVert w_i - g \rVert_2 \le d_{\text{horizon}}, \\
1, & \text{otherwise},
\end{cases}
\end{equation}

To evaluate the critic at arbitrary points along sampled trajectories, we construct critic observations by holding non-terrain inputs (e.g., speed and waypoint spacing) fixed while varying only the lookahead terrain observation. This isolates the contribution of terrain geometry to the critic’s value estimates, ensuring that differences in residual cost across trajectories arise from terrain variation rather than changes in control inputs or waypoint configuration.

\begin{figure}[!t]
    \vspace{2mm}
    \centering
    \includegraphics[width=0.89\linewidth]{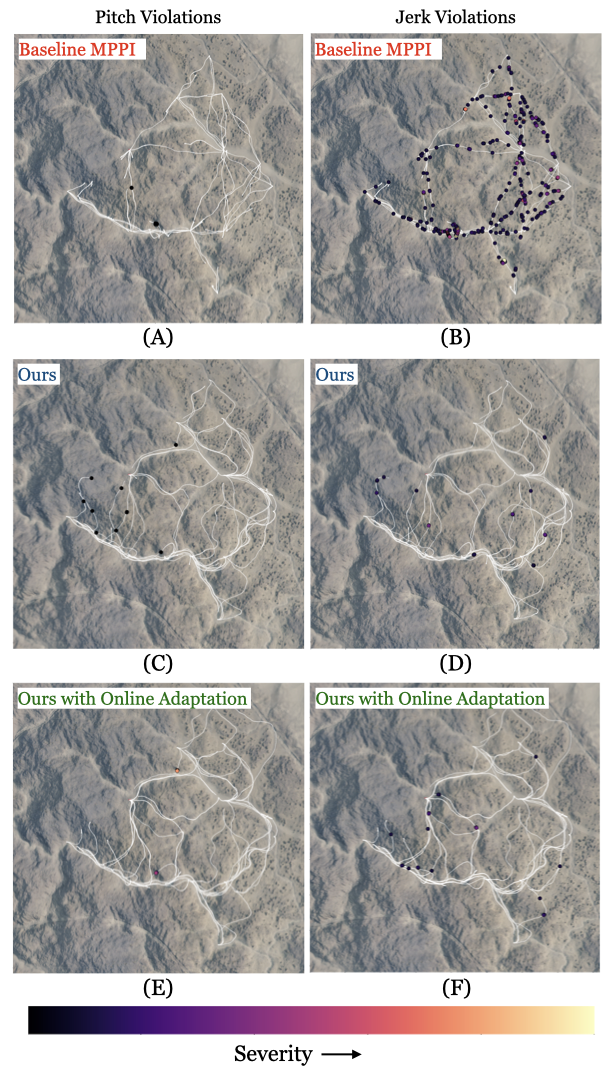}
    \caption{Comparison of pitch and vertical jerk violations across all 64 evaluation trajectories for the MPPI baseline ((A), (B)), our algorithm ((C), (D)), and our algorithm with online adaptation ((E), (F)). Panels (A), (C), and (E) show pitch violations exceeding $0.4$ rad ($22.9^\circ$), while (B), (D), and (F) show vertical jerk violations exceeding $10 \text{e}^{-2}$ m/$\text{s}^3$. The violations are colored by their severity and are no more than 0.7 rad. and 2 m/s$^3$ for pitch and jerk respectively. The MPPI baseline includes an explicit pitch cost term, whereas our method is trained with a jerk penalty in the reward function. When divided into equally-spaced buckets of low, medium, and high violation severity, the MPPI baseline recorded 317, 51, and 19 jerk violations and 84, 0, and 0 pitch violations, respectively; ours recorded 22, 1, and 0 jerk violations and 11, 1, and 0 pitch violations; and ours with online adaptation recorded 31, 0, and 0 jerk violations and 1, 1, and 3 pitch violations.
 }
    \label{fig:pitch_jerk_comparson}
\end{figure}

Lookahead terrain observations are obtained by sampling elevation patches from a nadir-pointing camera in the simulator centered at each waypoint along the trajectory. The planning horizon is limited to 40 m to reflect the range of typical onboard mapping systems, which can reliably provide local elevation maps up to 50 m or further ~\cite{roadrunner, tartandrive2.0}. Notably, our method does not require access to a global map or semantic cost annotations.

\subsection{Policy Evaluation for Critic Fine-Tuning and Continual Learning}\label{sec:methods:pe}
The pre-trained critic from Section \ref{sec:methods:ppo} is further refined through an offline policy evaluation step. Due to the large, sparse, negative penalties in the reward function, the critic severely underestimates value, although it appropriately learns to estimate \textit{relative} value correctly (for example, smooth terrain is estimated to have higher value than uneven terrain, but the values of both are underestimated).

To recalibrate the critic, rollout data is collected by executing the pre-trained policy, pre-trained critic, and MPPI planner across all training trajectories to record observation-reward pairs. Monte Carlo returns computed from these rollouts are then used to fine-tune the critic, reducing the gap between predicted value and actual return. The critic is trained for 1000 epochs on this dataset. The resulting critic maintains its ability to distinguish terrain difficulty while providing better-calibrated value estimates and is used in all subsequent experiments, both with and without online adaptation.

To enable continual learning from driving experience, we incorporate an online adaptation mechanism. Every $n_{\text{update freq.}} = 10$ control steps, newly collected rollout data is added to a replay buffer, and the value function is fine-tuned on the full replay buffer for 100 epochs. This online adaptation improves performance on unseen trajectories compared to the hierarchical algorithm without adaptation, as discussed in Section \ref{sec:experimental_results}.

\section{EXPERIMENTAL RESULTS}\label{sec:experimental_results}

We evaluate our algorithm through ablations and comparisons against baselines. We compare three algorithms: (i) our hierarchical algorithm without online adaptation, (ii) our hierarchical algorithm with online adaptation, and (iii) a baseline MPPI implementation.

The baseline MPPI implementation uses a kinematic bicycle model (KBM) for forward rollouts and includes goal and pitch cost terms, which were implemented following \cite{han2024mppi}. We do not include additional cost terms from \cite{han2024mppi}, such as rollover and semantic costs. The semantic cost is excluded to ensure a fair comparison, as our algorithm does not use semantic information. Our algorithm is deployed as described in Section \ref{sec:methods}, using the offline fine-tuned critic from Section \ref{sec:methods:pe}. For the online variant, the critic is updated every $n_{\text{update freq.}} = 10$ control steps (2 seconds), and fine-tuned for 100 epochs at each update. All implementation details and training hyperparameters for both the baseline and our method are provided in the Appendix.

\begin{figure}[t]
    \vspace{2mm}
    \centering
    \includegraphics[width=0.95\linewidth]{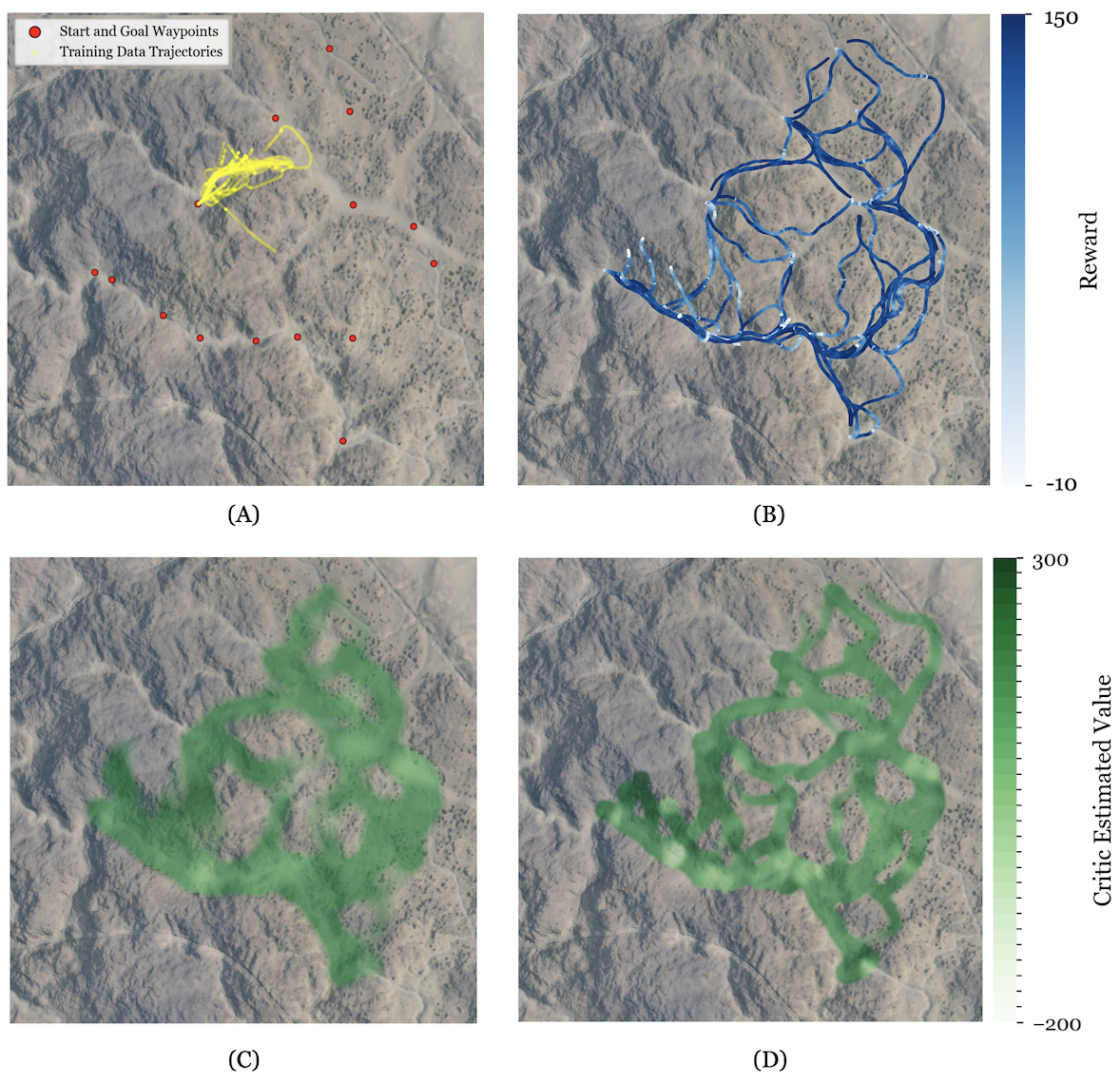}
    \caption{(A) Visualization of training data coverage and evaluation start–goal waypoints. (B) Trajectories traversed by the online adaptation variant of our algorithm, colored by the incurred reward as defined in Eq. \ref{eq:reward_fn}. (C) and (D) show value estimates of the traversed terrain during execution for our algorithm without and with online adaptation, respectively. These value maps can be interpreted as implicit cost maps over the visited regions. After online adaptation, regions associated with lower reward decrease in value (i.e., appear lighter), reflecting updated terrain estimates. Despite this, the algorithm continues to favor on-trail paths, as they are predicted to have higher value than off-trail terrain. All colorbars are scaled between the 10th and 90th percentiles of the plotted values.}
    \label{fig:cost_map}
\end{figure}

All algorithms are evaluated on 64 randomly generated start–goal pairs in the BeamNG Johnson Valley lakebed environment \cite{beamng}. Although trajectories may pass through terrain seen during training, each trajectory includes at least some unseen terrain, and all have either an unseen start or goal waypoint. (See Fig. \ref{fig:cost_map} (A) for a visualization of the training data coverage and start-goal evaluation waypoints.) The start–goal distances range from 70.0 m to 170.0 m to match the training distribution. Each episode is limited to 50 seconds (250 control steps at 5 Hz). All trajectories are theoretically feasible within this allotted time, and thus this limit provides a consistent time budget to evaluate traversal efficiency.

\subsection{Trajectory-Level Behavior and Terrain-Aware Navigation}

Fig.~\ref{fig:traj_comparison} shows a qualitative comparison of trajectories generated by baseline MPPI, our method, and our method with online adaptation across representative unseen start–goal pairs. These examples highlight consistent differences in how each method reasons about terrain and selects paths.

The baseline MPPI implementation tends to follow routes that remain close to the straight-line path between the start and goal, deviating only when necessary to reduce pitch cost. As a result, it frequently commits to traversing uneven or difficult terrain, which can lead to failure. This behavior is evident in Fig.~\ref{fig:traj_comparison}(A), (B), and (E), where MPPI becomes stuck and is unable to complete the route.

In contrast, our methods produce trajectories that better navigate difficult terrain. It often selects paths that deviate from the straight-line direction in order to avoid rough regions, even without explicit terrain labels. For example, in Fig.~\ref{fig:traj_comparison}(C), our method follows an existing trail rather than cutting directly across uneven terrain, indicating that the learned value function captures preferences for safer, more traversable regions even when they are unseen.

The online adaptation variant further improves this behavior by adjusting the value function during execution. This enables the system to react to local terrain conditions that were not fully captured during offline training. In Fig.~\ref{fig:traj_comparison}(A), although the non-adaptive method incurs damage about halfway through the trajectory, the online adaptation variant avoids this region through online updates alone, without prior knowledge that it is unsafe (this can be seen in the region where the blue and green lines deviate in Fig.~\ref{fig:traj_comparison}(A)). More broadly, the online method produces trajectories that more consistently avoid high-cost terrain and exhibit larger deviations when necessary. However, this adaptation is not always sufficient to guarantee success, as shown in Fig.~\ref{fig:traj_comparison}(D), where the method reaches the time limit without completing the trajectory.

Overall, these examples illustrate that our approach, particularly with online adaptation, leads to safer and more terrain-aware navigation compared to baseline MPPI, while maintaining the ability to generalize to unseen environments.

\begin{table}[t]
\vspace{3mm}
\centering
\small
\renewcommand{\arraystretch}{0.7}
\begin{tabular}{lccc}
\toprule
\textbf{Metric} & \textbf{MPPI} & \textbf{Ours} & \textbf{Ours Online} \\
\midrule

\multicolumn{4}{l}{\textit{Task Performance}} \\
Success (sum) & 35 (55\%) & \textbf{55 (86\%)} & 52 (81\%) \\
Failure & 29 (45\%) & 9 (14\%) & 12 (19\%) \\
\quad Stuck & 28 & 7 & 6 \\
\quad Timeout & 1 & 2 & 6 \\
Dist. Traveled (m) & 5495 & 9252 & $1.29\times10^{5}$ \\
Displacement (m) & 5157 & 6966 & 6813 \\

\midrule

\multicolumn{4}{l}{\textit{Z-axis Motion} ($\times 10^{2}$)} \\
$z$-jerk p10 & 0.17 & 0.14 & \textbf{0.12} \\
$z$-jerk p90 & 6.60 & 2.75 & \textbf{2.60} \\
$z$-jerk mean & 2.9 $\pm$ 5.9 & 1.27 $\pm$ 1.4 & \textbf{1.15 $\pm$ 1.3} \\

$z$-accel p10 & 1.2 & 0.6 & \textbf{0.4} \\
$z$-accel p90 & 32 & 12 & \textbf{11} \\
$z$-accel mean & 14 $\pm$ 17 & 5.4 $\pm$ 5.9 & \textbf{4.8 $\pm$ 5.4} \\

\midrule
\multicolumn{4}{l}{\textit{Orientation (rad.)} ($\times 10^{2}$)} \\
Pitch p10 & 2.0 & \textbf{1.0} & \textbf{1.0} \\
Pitch p90 & 27 & \textbf{20} & 21 \\
Pitch mean & 13 $\pm$ 9.9 & \textbf{8.0 $\pm$ 7.6} & \textbf{8.0 $\pm$ 7.9} \\

Roll p10 & 2.0 & \textbf{1.0} & \textbf{1.0} \\
Roll p90 & 26 & 19 & \textbf{18} \\
Roll mean & 12 $\pm$ 9.1 & \textbf{8.0 $\pm$ 7.9} & 9.0 $\pm$ 7.0 \\

\bottomrule
\end{tabular}

\caption{Comparison of navigation performance and terrain interaction metrics across methods. Each algorithm is evaluated on 64 randomly sampled start–goal pairs, all unseen during training. Episodes terminate after a maximum of 50 seconds (250 control steps) and are marked as failures if the goal is not reached within this budget (denoted ``timeout"). Other episodes terminate with the vehicle stuck (on a steep slope, for example), and these failures are denoted ``stuck." Reported values reflect trajectory statistics aggregated over all trials; lower values in Z-axis motion and orientation indicate smoother traversal. Note: p10 and p90 represent the 10th and 90th percentile of values, respectively.}
\label{table:results}
\end{table}

\subsection{Quantitative Evaluation of Traversal Performance}

\subsubsection{High-level Performance Metrics and Failure Modes}

Table~\ref{table:results} summarizes performance across all 64 evaluation trajectories. Our method and its online variant succeed on 55 and 54 trajectories, respectively, compared to 35 for the MPPI baseline, demonstrating improved robustness in challenging off-road terrain. We also report total distance traveled and displacement to quantify how much each method deviates from straight-line paths. Our methods exhibit significantly higher distance-to-displacement ratios than the MPPI baseline, indicating less direct trajectories. This behavior reflects the value function’s preference for smoother and safer terrain, and corresponds to reduced vertical jerk, acceleration, and orientation deviations (quantified in Table~\ref{table:results}), indicating more stable traversal. The online adaptation variant further amplifies this effect, often producing longer, more conservative trajectories. While this can improve terrain avoidance, it also leads to increased path lengths and occasional failures due to exceeding the 250-step time budget. Additional failure modes across all methods include getting stuck on steep slopes and rocks.

\subsubsection{Traversal Disturbance Metrics}
Our method achieves lower peak and average vertical jerk and acceleration compared to both the baseline MPPI and the non-adaptive variant, indicating that the planned trajectories better account for terrain-induced disturbances and that our controller handles uneven terrain in a way that reduces disturbances. These improvements are consistent across both lower (p10) and upper (p90) percentiles, suggesting that our approach reduces not only typical but also worst-case disturbances. Similarly, improvements in pitch and roll statistics show that our method more effectively avoids geometrically challenging terrain, resulting in smoother and more stable vehicle motion. The online adaptation variant further improves these metrics in most cases, achieving the lowest jerk and acceleration overall. These gains come alongside higher success rates, demonstrating that our approach improves both traversal safety and task completion, even though it follows less direct paths. Figure \ref{fig:pitch_jerk_comparson} shows trajectories traversed and locations of pitch and jerk violations for each algorithm.

\subsubsection{Implicit Cost Map Visualization}
We visualize the algorithm’s estimated value function along the executed trajectories as a heatmap over the traversed terrain in Fig. \ref{fig:cost_map}. This provides an interpretable view of the implicit cost map encoded by the learned value function. Regions predicted to have higher value correspond to smoother and more traversable terrain, while rough or disruptive areas are assigned lower value. As expected, the value estimates align with terrain quality, indicating that the learned model captures meaningful structure in the environment.

\section{CONCLUSIONS \& FUTURE WORK}
\raggedbottom

In this paper, we introduced a hierarchical off-road navigation framework in which an MPPI-style planner uses the value function of a low-level RL controller as an execution-aware cost signal. This coupling enables the planner to select trajectories that are both goal-directed and realizable under the controller’s true behavior, without requiring explicit semantic cost maps. We also proposed an online policy evaluation procedure that adapts the value function during deployment, improving robustness on unseen terrain. Our results in high-fidelity simulation show that value-guided planning reduces vehicle disturbance and improves success rates compared to baseline MPPI.

Future work will incorporate visual information to extend the value function beyond geometric terrain hazards to challenging terrain properties as well. While this work focuses on terrain geometries such as hills, ditches, and uneven surfaces, deployment in densely cluttered environments will require obstacle detection. We therefore plan to augment MPPI with a lethal obstacle cost for trajectories intersecting detected obstacles, while retaining the learned value function to evaluate controller-dependent terrain difficulty. Finally, we plan to validate the framework on a physical vehicle.




\FloatBarrier
\section*{APPENDIX}

\begin{table}[H]
\centering

\vspace{0.5em}
\small
\begin{tabular}{lc}
\midrule
\textbf{Hyperparameter} & \textbf{Value} \\
\midrule

\multicolumn{2}{l}{\textit{PPO Policy Training}} \\
Training data coverage (m) & $3.5\times 10^{3}$ \\
Waypoint spacing (m) & 4 \\
Waypoint acceptance thresh. (m) & 3 \\
Reward weight $\beta_1$ & 1000 \\
Reward weight $\beta_2$ & -1000 \\
Reward weight $\beta_3$ & -100 \\
Reward weight $\beta_4$ & -5 \\
\midrule

\multicolumn{2}{l}{\textit{MPPI-style Planner (Ours)}} \\
\# Trajectory samples & 100 \\
Horizon (m) & 40 \\
Waypoint spacing (m) & 4 \\
$\lambda_{\text{goal}}$ & $1 \times 10^{-3}$ \\
$\lambda_{\text{res}}$ & 1 \\
\midrule

\multicolumn{2}{l}{\textit{Offline Finetuning}} \\
Training data coverage (m) & $3.5\times 10^{3}$ \\
\# Epochs & 1000 \\
Batch size & 256 \\
\midrule

\multicolumn{2}{l}{\textit{Online Adaptation}} \\
Update frequency (s) & 2 \\
\# Epochs & 100 \\
Batch size & 256 \\
\midrule

\multicolumn{2}{l}{\textit{MPPI Baseline}} \\
Pitch cost weight & 0.5 \\
Goal cost weight & 1.0 \\
\bottomrule
\end{tabular}
\caption{Hyperparameters used in all implementations. PPO policy training is performed using the standard Stable-Baselines3 implementation~\cite{sb3}.}
\label{table:hyperparameters}
\end{table}


\bibliographystyle{plain}
\bibliography{references}

\end{document}